\pdfoutput=1

\documentclass[11pt]{article}

\usepackage{EMNLP2022}

\usepackage{times}
\usepackage{latexsym}

\usepackage[T1]{fontenc}

\usepackage[utf8]{inputenc}

\usepackage{microtype}
\usepackage{xspace}
\usepackage{inconsolata}

\usepackage{microtype}
\usepackage{amsmath}
\usepackage{graphicx}
\usepackage{tabularx}
\usepackage{xspace}
\usepackage[draft]{todo}
\usepackage{stfloats}
\usepackage[group-separator={,},group-minimum-digits={3}]{siunitx}
\newcommand{\src}{\ensuremath{\mathbf{f}}} 
\newcommand{\trg}{\ensuremath{\mathbf{e}}} 
\newcommand{\simsrc}{\ensuremath{\mathbf{\tilde{f}}}} 
\newcommand{\initrg}{\ensuremath{\mathbf{e}^{\perp}}} 
\newcommand{\paratrg}{\ensuremath{\mathbf{e}^{\parallel}}} 
\renewcommand{\initrg}{\ensuremath{\mathbf{e}^{\neq}}} 
\renewcommand{\paratrg}{\ensuremath{\mathbf{e}^{\equiv}}} 
\renewcommand{\initrg}{\ensuremath{\mathbf{e}^{\sharp}}} 
\renewcommand{\paratrg}{\ensuremath{\mathbf{e}^{}}} 
\renewcommand{\initrg}{\ensuremath{\mathbf{\bar{e}}}} 
\renewcommand{\initrg}{\ensuremath{\mathbf{\tilde{e}}}} 
\newcommand{\editmt}[0]{Edit-MT\xspace{}}
\newcommand{\editlevt}[0]{Edit-LevT\xspace{}}
\newcommand{\bisync}[0]{Bi-sync\xspace{}}
\newcommand{\stkout}[1]{}

\title{Bilingual Synchronization: Restoring Translational Relationships with Editing Operations}

\author{Jitao Xu$^{\dag}$\ $\quad\quad\quad\quad$ Josep Crego$^{\ddag}$\ $\quad\quad\quad\quad$  François Yvon$^{\dag}$ \\ \\
$^\dag$Universit\'e Paris-Saclay, CNRS, LISN, 91400, Orsay, France \\
$^\ddag$SYSTRAN, 5 rue Feydeau, 75002, Paris, France \\
\texttt{\{jitao.xu,francois.yvon\}@limsi.fr, josep.crego@systrangroup.com} \\}

\begin{document}
\maketitle

\begin{abstract}
Machine Translation (MT) is usually viewed as a one-shot process that generates the target language equivalent of some source text from scratch. We consider here a more general setting which assumes an initial target sequence, that must be transformed into a valid translation of the source, thereby restoring parallelism between source and target. For this \emph{bilingual synchronization task}, we consider several architectures (both autoregressive and non-autoregressive) and training regimes, and experiment with multiple practical settings such as simulated interactive MT, translating with Translation Memory (TM) and TM cleaning. Our results suggest that \emph{one single generic edit-based} system, once fine-tuned, can compare with, or even outperform, dedicated systems specifically trained for these tasks.
\end{abstract}

\section{Introduction\label{sec:intro}}

Neural Machine Translation (NMT) systems have made tangible progress in recent years \citep{Bahdanau15neural,Vaswani17attention}, as they started to produce usable translations in production environments. NMT is generally viewed as a one-shot activity process in autoregressive approaches, which generates the target translation based on the sole source side input. Recently, Non-autoregressive Machine Translation (NAT) models have proposed to perform iterative refinement decoding \citep{Lee18deterministic,Ghazvininejad19mask,Gu19levenshtein}, where translations are generated through an iterative revision process, starting with a possibly empty initial hypothesis.

This paper focuses on the revision part of the machine translation (MT) process and consider \emph{bilingual synchronization} (\bisync), which we define as follows: given a pair of a source (\src) and a target (\initrg) sentences, \emph{which may or may not be mutual translations}, the task is to compute a revised version \paratrg{} of \initrg, such that \trg{} is an actual translation of \src. This is necessary when the source side of an existing translation is edited, requiring to update the target and keep both sides synchronized. \bisync{} subsumes standard MT, where the synchronization starts with an empty target (\initrg{} = []). Other interesting cases occur when parts of the initial target can be reused, so that the synchronization only requires a few changes.

\bisync{} encompasses several tasks: synchronization is needed in interactive MT (IMT, \citealp{Knowles16interactive}) and bilingual editing \citep{Bronner12cosyne}, with \initrg{} the translation of a previous version of \src; in MT with lexical constraints \citep{Hokamp17lexically}, where \initrg{} contains target-side constraints \citep{Susanto20lexically,Xu21editor}; in Translation Memory (TM) based approaches \citep{Bulte19fuzzy}, where \initrg{} is a TM match for a similar example; in automatic post-editing (APE) \citep{doCarmo21review}, where \initrg{} is an MT output.

We consider here several implementations of sequence-to-sequence models dedicated to these situations, contrasting an autoregressive model with a non-autoregressive approach. The former is similar to \citet{Bulte19fuzzy}, where the source sentence and the initial translation are concatenated as one input sequence; the latter uses the Levenshtein Transformer (LevT) of \citet{Gu19levenshtein}. We also study various ways to generate appropriate training samples (\src, \initrg, \paratrg). Our experiments consider several tasks, including TM cleaning, which attempts to fix and synchronize noisy segments in a parallel corpus. This setting is more difficult than \bisync, as many initial translations are already correct and need to be left unchanged. Our results suggest that one single AR system, once fine-tuned, can favorably compare with dedicated systems for each of these tasks. To recap, our main contributions are (a) the generalization of several tasks subsumed by a generic synchronization objective, allowing us to develop a unified perspective about otherwise unrelated subdomains of MT; (b) the design of a training procedure for a generic edit-based model; (c) an empirical validation on five settings and domains.

\begin{figure*}[ht]
  \center
  \includegraphics[width=\textwidth]{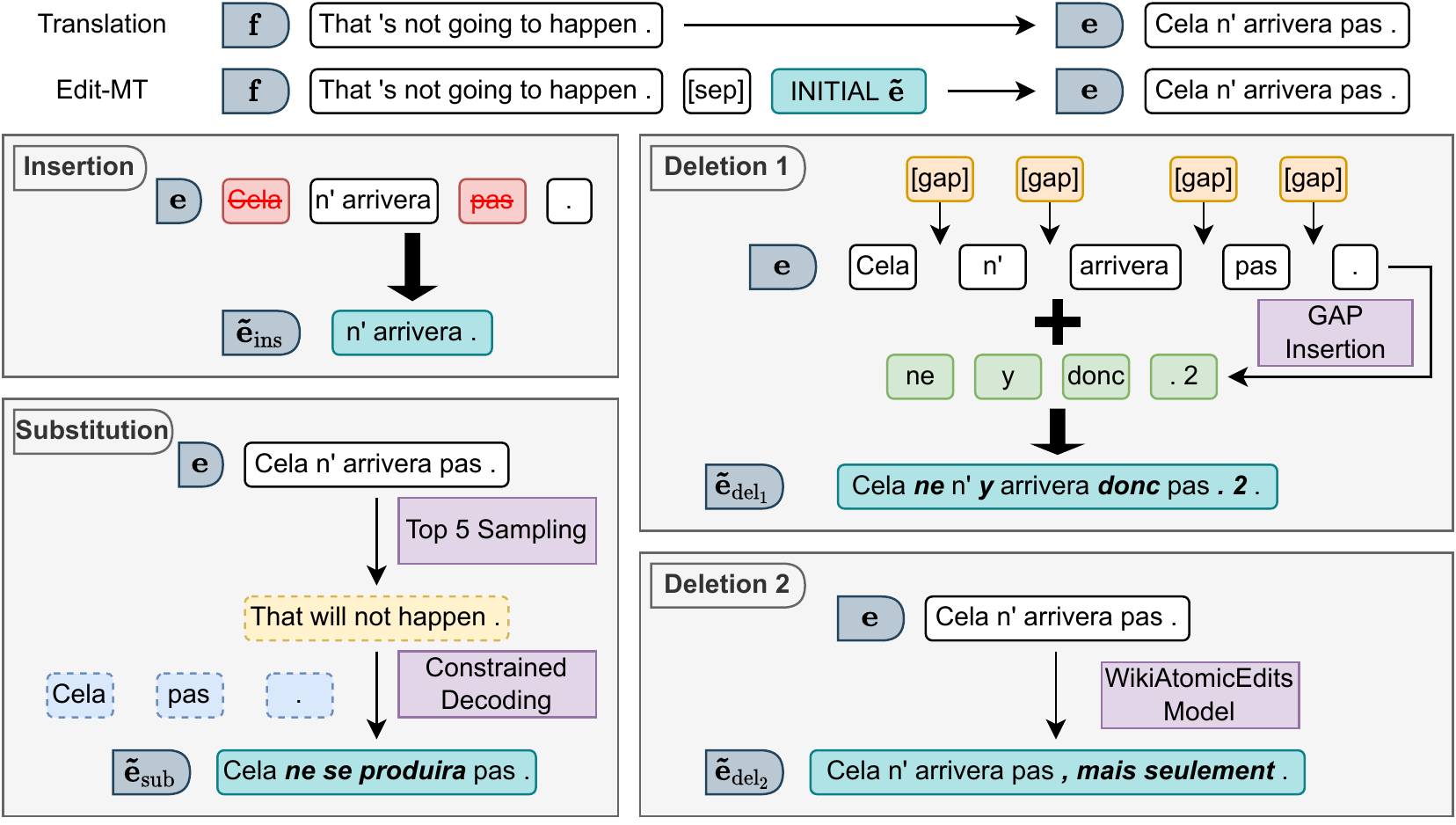}
  \caption{Methods for generating synthetic initial translations \initrg{} for each edit type. Rectangle purple boxes refer to separate models used to generate the desired operations. Differences in artificial initial translations (in blue boxes) are marked in bold. Initial translations $\initrg_{\operatorname{ins}}$ for insertion are generated by randomly removing segments in the reference sentence $\paratrg$. For $\initrg_{\operatorname{sub}}$, $\paratrg$ is first back-translated into an intermediate sentence $\src^*$ using top-$5$ sampling, then translated back to $\initrg_{\operatorname{sub}}$ with LCD. The first method to generate $\initrg_{\operatorname{del}}$ randomly inserts \texttt{[gap]} tokens into $\trg$ and decodes with a GAP insertion model \citep{Xiao22bitiimt}. $\initrg_{\operatorname{del}_1}$ is obtained by replacing \texttt{[gap]} with the predicted segments. The second method automatically edits $\trg$ with a model trained on WikiAtomicEdits data. \label{fig:datagen}}
\end{figure*}

\section{Methods\label{sec:method}}

\subsection{Generating Editing Data\label{ssec:data-generation}}

We consider a general scenario where, given a pair of sentences \src{} and \initrg, assumed to be related, but not necessarily parallel, we aim to generate a target sentence \paratrg{} that is parallel to \src. We would also like \initrg{} and \paratrg{} to be close, as \initrg{} is often a valid translation of a sentence \simsrc{} that is close to \src. Training such models requires triplets (\src, \initrg, \paratrg). While large amounts of parallel bilingual data are available for many language pairs, they are hardly ever associated to related translations \initrg{} (except for APE). We therefore study ways to simulate synthetic \initrg{} from \paratrg, while preserving large portions of \paratrg{} in \initrg. String edits can be decomposed into a sequence of three basic edits (insertions, substitutions and deletions), we design our artificial samples so that edits from \initrg{} to \paratrg{} only involve one type of operation (Figure~\ref{fig:datagen}). 

\paragraph{Insertions\label{para:ins}}

We mainly follow \citet{Xiao22bitiimt} to generate initial translations $\initrg_{\operatorname{ins}}$ for insertion by randomly deleting segments from \paratrg. For each \paratrg, we first randomly sample an integer $k$$\in$$[1,5]$, then randomly remove $k$ non-overlapping segments from \paratrg. The length of each removed segment is also randomly sampled with a maximum of $5$ tokens. We also impose that the overall ratio of removed segments does not exceed $0.5$ of $\trg$. Different from \citet{Xiao22bitiimt}, $\initrg_{\operatorname{ins}}$ does not include any placeholders to locate the positions of removed segments. This makes $\initrg_{\operatorname{ins}}$ a more realistic starting point as the insertion positions are rarely known in practical settings. Our preliminary experiments also show that identifying insertion positions makes the infilling task easier than when they are unknown.

\paragraph{Substitutions\label{para:sub}}

To simulate substitutions, we apply round-trip translation with lexically constrained decoding (LCD, \citealp{Post18fast}) to generate initial translations for substitution $\initrg_{\operatorname{sub}}$. Round-trip translation is already used for the APE task in \citet{Junczys-dowmunt16loglinear}. This requires two standard NMT models separately trained on parallel data, one for each direction. For each training example (\src, \paratrg), we first (a) translate $\paratrg$ into an intermediate source sentence $\src^*$ using top-$5$ sampling \citep{Edunov18understanding};\footnote{Early experiments showed that using sampling instead of beam search increases the diversity of the generated $\initrg_{\operatorname{sub}}$.} (b) generate an abbreviated version $\initrg'_{\operatorname{ins}}$ using the method described above for insertions. We then translate $\src^*$ using LCD, with $\initrg'_{\operatorname{ins}}$ as constraints, to obtain $\initrg_{\operatorname{sub}}$. In this way, we ensure that at least half of \paratrg{} remains unchanged in $\initrg_{\operatorname{sub}}$, while the other parts have been substituted. To increase diversity, $\initrg'_{\operatorname{ins}}$ (used to create $\initrg_{\operatorname{sub}}$) is sampled with a different random seed than $\initrg_{\operatorname{ins}}$ (used for the insertion task).

\paragraph{Deletions\label{para:del}}

Simulating deletions requires the initial translation $\initrg_{\operatorname{del}}$ to be an extension of $\paratrg$. We propose two strategies to generate $\initrg_{\operatorname{del}}$. The first uses a GAP insertion model as in \citet{Xiao22bitiimt}, in which word segments are randomly replaced with a placeholder $\texttt{[gap]}$ to generate $\initrg_{\operatorname{gap}}$. The task is then to predict the missing segments based on the concatenation of $\src$ and $\initrg_{\operatorname{gap}}$ as input. This differs from our own insertion task, as (a) insertion positions are identified as a \texttt{[gap]} symbol in $\initrg_{\operatorname{gap}}$ and (b) generation only computes the sequence of missing segments $\trg_{\operatorname{seg}}$, rather than a complete sentence. 

We use GAP to generate extra segments for a pair of parallel sentences as follows. We randomly insert $k$$\in$$[1,5]$ \texttt{[gap]} tokens into $\paratrg$, concatenate it with $\src$ and use GAP to predict extra segments, yielding the synthetic target sentence $\initrg_{\operatorname{del}_1}$. This method always extends parallel sentences with additional segments on the target side. However, these segments are arbitrary and may not contain any valid semantic information, nor be syntactically correct. 

We thus consider a second strategy, based on actual edit operations collected in the WikiAtomicEdits dataset\footnote{\url{https://github.com/google-research-datasets/wiki-atomic-edits}} \citep{Faruqui18wikiatomicedits}, which contains edits of an original segment $\mathbf{x}$ and the resulting segment $\mathbf{x}^{\prime}$, with exactly one insertion or deletion operation for each example, collected from Wikipedia edit history. This notably ensures that both versions of each utterance are syntactically correct. We treat the deletion data of WikiAtomicEdits as ``reversed'' insertions, and use both of them to train a seq-to-seq \texttt{wiki} model ($\mathbf{x}_{\operatorname{short}}$$\to$$\mathbf{x}_{\operatorname{long}}$), generating longer sentences from shorter ones. The \texttt{wiki} model is then used to expand $\trg$ into an $\initrg_{\operatorname{del}_2}$. Compared to $\initrg_{\operatorname{del}_1}$, $\initrg_{\operatorname{del}_2}$ is syntactically more correct. However, it is also by design very close (one edit away) to $\paratrg$.

As both simulation methods have merits and flaws, we randomly select examples from $\initrg_{\operatorname{del}_1}$ and $\initrg_{\operatorname{del}_2}$ to build the final synthetic initial translation samples for the deletion operation $\initrg_{\operatorname{del}}$.

\paragraph{Copy and Translate Operations}

To handle parallel sentences that do not require any changes, we add a fourth copy operation, where the initial translation $\initrg_{\operatorname{cp}}$ is equal to the target sentence ($\initrg_{\operatorname{cp}}$$=$$\paratrg$). Hence, the data used to learn edit operations is built with triplets (\src, \initrg, \paratrg) where $\initrg$ is uniformly randomly selected from $\initrg_{\operatorname{ins}}$, $\initrg_{\operatorname{sub}}$, $\initrg_{\operatorname{del}}$ and $\initrg_{\operatorname{cp}}$. Finally, to maintain the capacity to perform standard MT from scratch, we also consider samples where \initrg{} is empty. The implementation of standard MT varies slightly upon approaches, as we explain below.

\subsection{Model Architectures\label{ssec:model}}

We implement \bisync{} with Transformer-based \citep{Vaswani17attention} autoregressive and non-autoregressive models. The former (\editmt) is a regular Transformer with a combined input made of the concatenation of $\src$ and $\initrg$; the latter (\editlevt) is the LevT of \citet{Gu19levenshtein}.

\paragraph{\editmt}

In this model, $\initrg$ is simply concatenated to $\src$, with a special token to separate the two sentences. This technique has been used \textit{e.g.}\ in \citet{Dabre17enabling} for multi-source MT or in \citet{Bulte19fuzzy} for translating with a similar example. The input side of the editing training data is thus \src{} \texttt{[sep]} \initrg, as shown in Figure~\ref{fig:datagen} (top). 

On the target side, we add a categorical prefix to indicate the type of edit(s) associated with a given training sample, as is commonly done for multi-domain or multilingual MT. For each basic edit (insertion, substitution and deletion), we use a binary tag to indicate if the operation is required. For instance, an $\initrg_{\operatorname{ins}}$ needing insertions would have tags \texttt{[ins]} \texttt{[!sub]} \texttt{[!del]} prepended to $\paratrg$. Copy corresponds to all three tags set to negative as \texttt{[!ins]} \texttt{[!sub]} \texttt{[!del]}. The tagging scheme provides us with various ways to perform edit-based MT:
(a) we can perform inference without knowing the required edit type of $\initrg$ by generating tags then translations;
(b) when the edits are known, we can generate translations with desired edits by using the corresponding tags as a forced prefix;
(c) inference can also only output the edit tags and predict the relation between $\src$ and $\initrg$.
The ability to perform standard MT is preserved by training with a balanced mixture of editing data and parallel data. The latter corresponds to an empty $\initrg$. For these examples, the target side does not contain any tags.

\paragraph{\editlevt}

LevT is trained with a randomly noised version of the reference as initial target, and decodes with empty sentences. In \bisync, we instead initialize the target side with the given $\initrg$ for training and inference. To perform standard MT with the same model, we train with a tunable mixture of these two strategies, where $p$ controls the proportion of each type of samples. Taking $p=0$ is equivalent to train a LevT model with only parallel data. We use $p=0.5$ in our experiments, making it equivalent to mixing editing and parallel data for the \editmt{} model. The value of $p$ can be carefully designed with a schedule or curriculum to optimize the behavior of \editlevt, which we leave for future work. For \editlevt, we do not use any tags, as \editlevt{} already includes an internal mechanism to predict the edit operation(s).

\section{Bilingual Re-synchronization\label{sec:bisynch}}

\subsection{Datasets and Experimental Settings\label{ssec:data-bisynch}}

We first evaluate \editmt{} and \editlevt{} on a basic resynchronization task where \initrg{} is assumed to be the translation of a former version of \src, and only a limited amount of edits is sufficient to restore parallelism. We conduct experiments on WMT14 English-French data\footnote{\url{https://www.statmt.org/wmt14}} in both directions (En-Fr \& Fr-En) and evaluate on two test sets. The first is an artificial derivation of the standard newstest2014 set, and the second is the small parallel sentence compression dataset\footnote{\url{https://github.com/fyvo/ParallelCompression}} of \citet{Ive16parallel}.

For the original newstest2014, we generate an initial version $\initrg$ for each test sentence and each edit operation according to the methods of Section~\ref{ssec:data-generation}. For deletion, we test the performance of both generation methods, resulting in four versions (Ins, Sub, Del$_1$, Del$_2$) of newstest2014 with \num{3003} sentences each. The sentence compression dataset contains a subset of documents also selected from newstest2014, where sentences are compressed by human annotators while remaining parallel in the two languages. We only retain utterances for which the compressed and original versions actually differ on both sides, resulting in $526$ test sentences.

Both experiments use the same training data, where we discard examples with invalid language tag as computed by \texttt{fasttext} language identification model\footnote{\url{https://fasttext.cc/}} \citep{Bojanowski17enriching}, yielding a training corpus of $33.9$M examples. We tokenize all data using Moses\footnote{\url{https://github.com/moses-smt/mosesdecoder}} and build a shared source-target vocabulary with $32$K Byte Pair Encoding (BPE) units \citep{Sennrich16BPE} learned with \texttt{subword-nmt}.\footnote{\url{https://github.com/rsennrich/subword-nmt}} Since we use both parallel and artificial editing data to train edit-based models, the total training data contains about 68M utterances.

We conduct experiments using \texttt{fairseq}\footnote{\url{https://github.com/pytorch/fairseq}} \citep{Ott19fairseq}. \editmt{} relies on the Transformer-base model of \citet{Vaswani17attention}. Model and training configurations are in Appendix~\ref{sec:bisynch-appendix}. Performance is computed with SacreBLEU \citep{Post18sacrebleu}.

\subsection{Results\label{ssec:bisynch-result}}

We first separately evaluate the learnability of each edit operation on our synthetic newstest2014 sets. We also derive two tasks from the compression dataset: parallel sentence compression (comp) and extension (ext). For compression, the task consists of producing a compressed target sentence $\trg_{\operatorname{comp}}$ given the compressed source $\src_{\operatorname{comp}}$ and the original target $\trg$. For extension, the model should produce $\paratrg$ with $\src$ and the compressed target $\trg_{\operatorname{comp}}$. These two tasks are respectively similar to the deletion and insertion tasks. There are slight differences, though, as (a) \initrg{} for these settings is always syntactically correct, (b) segments that are removed or inserted are selected for their lower informativeness. Therefore, these tasks are more about restoring an adequate, rather than a fluid, translation. 

As mentioned in Section~\ref{ssec:model}, the generation of translations in \editmt{} models is conditioned on predicted or oracle editing tags that are prefixed to the output: these two situations are contrasted using forced-prefix decoding with the correct tags. For the compression and extension tasks, we use the deletion and insertion tags, respectively. 

\begin{table*}[!ht]
  \center
  \scalebox{0.85}{
  \begin{tabular}{l|cccccccccc|c}
  \hline
  En-Fr & 0 & 1 & 2 & 3 & 4 & 5 & 6 & 7 & 8-10 & >10 & All \\
  \hline
  $N$ & 458 & 2076 & 1133 & 1152 & 1117 & 1033 & 880 & 776 & 1852 & 1535 & 12012 \\
  \hline
  copy & 100.0 & 88.9 & 84.7 & 81.5 & 77.8 & 75.2 & 70.3 & 68.0 & 65.3 & 52.9 & 75.9 \\
  \hline
  \editmt & \textbf{96.4} & \textbf{92.6} & 89.3 & 87.3 & 85.8 & 83.3 & 82.3 & 80.4 & 79.4 & 71.7 & 84.1 \\
  $\quad$+ tag & 95.5 & \textbf{92.6} & \textbf{90.4} & \textbf{88.3} & \textbf{86.9} & \textbf{84.4} & \textbf{83.0} & \textbf{81.3} & \textbf{80.1} & \textbf{72.1} & \textbf{84.7} \\
  \editlevt & 95.9 & 88.3 & 84.3 & 80.6 & 77.2 & 74.2 & 69.7 & 67.9 & 65.3 & 57.8 & 74.8 \\
  \hline
  \hline
  Fr-En & 0 & 1 & 2 & 3 & 4 & 5 & 6 & 7 & 8-10 & >10 & All \\
  \hline
  $N$ & 341 & 2050 & 1439 & 1178 & 1125 & 1054 & 945 & 832 & 1643 & 1405 & 12012 \\
  \hline
  copy & 100.0 & 90.1 & 85.8 & 80.6 & 76.4 & 71.8 & 68.7 & 67.0 & 63.8 & 51.7 & 75.6 \\
  \hline
  \editmt & 97.6 & 92.8 & 91.1 & 87.8 & 83.8 & 82.5 & 80.1 & 80.2 & 77.3 & 69.2 & 83.6 \\
  $\quad$+ tag & \textbf{98.3} & \textbf{93.2} & \textbf{91.6} & \textbf{88.6} & \textbf{84.8} & \textbf{83.3} & \textbf{80.8} & \textbf{80.5} & \textbf{77.9} & \textbf{69.9} & \textbf{84.2} \\
  \editlevt & 97.9 & 89.0 & 84.4 & 79.6 & 75.0 & 71.0 & 67.3 & 66.2 & 64.8 & 54.6 & 73.6 \\
  \hline
  \end{tabular}
  }
  \caption{BLEU scores for all edit types (Ins, Sub, Del$_1$, Del$_2$) broken down by the edit distance $\Delta$ between $\initrg$ and $\trg$ for both En-Fr and Fr-En. Each column represents a range of distances. $N$ denotes the number of sentences in each group. \textit{All} is computed by concatenating all test sentences.\label{tab:edit-dist-bisynch}}
\end{table*}

\begin{table}[!ht]
  \center
  \scalebox{0.85}{
  \begin{tabular}{l|cccc|cc}
  \hline
  Model & Ins & Sub & Del$_1$ & Del$_2$ & Comp & Ext \\
  \hline
  copy & 54.0 & 71.5 & 71.0 & 78.7 & 68.4 & 66.7 \\
  \hline
  \editmt & 75.9 & 77.0 & 86.9 & \textbf{94.7} & 73.1 & 67.9 \\
  $\quad$+ tag & \textbf{76.9} & \textbf{78.5} & \textbf{88.6} & \textbf{94.7} & \textbf{74.0} & \textbf{72.7} \\
  \editlevt & 65.3 & 73.9 & 72.5 & 78.7 & 67.8 & 67.7 \\
  \hline
  \end{tabular}
  }
  \caption{BLEU scores for \editmt{} and \editlevt{} on resynchronization tasks for En-Fr. Deletions are evaluated separately for two generation methods (Del$_1$ and Del$_2$). \textit{+ tag} refers to decoding with the oracle tag as a forced-prefix. Best performance is in bold. \label{tab:bisynch-enfr}}
\end{table}

\begin{table}[!ht]
  \center
  \scalebox{0.85}{
  \begin{tabular}{l|cccc|cc}
  \hline
  Model & Ins & Sub & Del$_1$ & Del$_2$ & Comp & Ext \\
  \hline
  copy & 51.8 & 70.9 & 71.0 & 78.7 & 63.4 & 61.4 \\
  \hline
  \editmt & 73.6 & 74.6 & 87.5 & 95.8 & 65.8 & 69.3 \\
  $\quad$+ tag & \textbf{74.6} & \textbf{76.2} & \textbf{89.1} & \textbf{96.2} & \textbf{67.0} & \textbf{71.6} \\
  \editlevt & 66.5 & 72.4 & 72.3 & 78.4 & 62.5 & 64.3 \\
  \hline
  \end{tabular}
  }
  \caption{BLEU scores for \editmt{} and \editlevt{} models on resynchronization tasks for Fr-En.\label{tab:bisynch-fren}}
\end{table}

Tables~\ref{tab:bisynch-enfr} and \ref{tab:bisynch-fren} report results for both directions, to be compared with a ``do-nothing'' baseline which simply copies $\initrg$ as the output. \editmt{} is able to edit the given $\initrg$ for all types of required edits much better than \editlevt. It obtains large gains over the \textit{copy} baseline for insertion,\footnote{BLEU gains for the insertion operation are artificially high. This is because the baseline is hindered by a high brevity penalty.} substitution and deletion for both directions. When tested on the compression and extension tasks, which have different edit distributions to the artificial editing data, \editmt{} still improves $\initrg$ by $1.2$-$4.7$ BLEU for En-Fr and $2.4$-$7.9$ BLEU for Fr-En. By prefixing \editmt{} with the oracle editing type tags, we can further boost the performance on almost every task for both directions. \editlevt{} can also improve $\initrg$ in most test situations, even though the gains are lower than \editmt{} models. However, due to the non-autoregressive nature of \editlevt{}, it obtains a decoding speedup of $2.3$$-$$3\times$ with respect to \editmt{} when tested with the same inference batch size on the same hardware, as recommended by \citet{Helcl22nonautoregressive}.

To better understand \editmt{} and \editlevt{} on the resynchronization task, we further analyze their performance with respect to the edit distance $\Delta$ between $\initrg$ and $\trg$. For the results in Table~\ref{tab:edit-dist-bisynch}, we merge test sentences of all edit types (Ins, Sub, Del$_1$, Del$_2$) into one test set, then break them down by the value of $\Delta$. For both directions, prefixing the oracle editing tags for \editmt{} yields a stable improvement for almost all $\Delta$. \editlevt{} performs similar to \editmt{} when no editions are needed, but only starts to improve the \textit{copy} baseline when more edits are required ($\Delta\geq8$).

\section{Translating with Translation Memories\label{sec:MTwithTM}}

As explained above, \bisync{} encompasses example-based MT, whereby an existing similar translation retrieved from a TM is turned into an adequate translation of the source. \editmt{} actually uses the same architecture as the retrieval-based models of \citet{Bulte19fuzzy,Xu20boosting}. In this section, we study the performance of our synchronization models in this practical scenario.

\subsection{Datasets\label{ssec:tm-data}}

We use the same multi-domain corpus as \citet{Xu20boosting}, which contains $11$ different domains for the En-Fr direction, collected from OPUS\footnote{\url{https://opus.nlpl.eu/}} \citep{Tiedemann12parallel}. We search for similar translations using Fuzzy Match.\footnote{\url{https://github.com/SYSTRAN/fuzzy-match}} The similarity between two English source sentences is computed as:
\begin{equation}
  \operatorname{sim}(\src, \simsrc) = 1 - \frac{\operatorname{ED}(\src, \simsrc)}{\max(|\src|, |\simsrc|)}, \label{eq:similarity}
\end{equation}
where $\operatorname{ED}(\src, \simsrc)$ is the edit distance between \src{} and \simsrc, and $|\src|$ is the length of \src. The intuition is that the closer \src{} and \simsrc{} is, the more suitable \initrg{} will be. To study the ability of our models to actually make use of TMs instead of memorizing training examples, we also test on two unseen domains: OpenOffice from OPUS and the PANACEA environment corpus\footnote{\url{http://catalog.elda.org/en-us/repository/browse/ELRA-W0057/}} (ENV). Detailed description and statistics about these corpora are in Appendix~\ref{sec:tm-appendix}.
\citet{Xu20boosting} proposed an alternative where, for each similar translation, they masked out segments that were not aligned with the source. This means that the initial similar translation only contains segments that are directly related to the source input. We also reproduce this \textit{related} setting, which is very similar to the insertion task.

As we are mostly interested in the edit behavior, we split the data by keeping \num{1000} sentences with a sufficiently similar translation ($\operatorname{sim}>0.6$) in the TM as the test set for each domain. The remaining data is used for training. We use all found similar translations for training and only the best similar translation for testing. This results in 4.4M parallel sentences (\textit{para}), in which 2.6M examples are also associated to a similar translation (\textit{similar}) or just the related segments (\textit{related}). Data preprocessing is similar as in Section~\ref{ssec:data-bisynch}.

\begin{table*}[!ht]
  \center
  \scalebox{0.85}{
  \begin{tabular}{l|ccccccccccc|c}
  \hline
  Model & ECB & EMEA & Epps & GNOME & JRC & KDE & News & PHP & TED & Ubuntu & Wiki & All \\
  \hline
  copy & 59.8 & 64.5 & 34.4 & 70.3 & 67.6 & 55.3 & 12.0 & 38.6 & 30.8 & 51.6 & 47.4 & 52.6 \\
  \hline
  \texttt{FM} & \textbf{72.1} & \textbf{72.3} & \textbf{58.3} & \textbf{80.6} & \textbf{83.2} & \textbf{66.9} & 28.0 & \textbf{47.2} & \textbf{62.9} & \textbf{69.3} & 68.8 & \textbf{67.3} \\
  \texttt{FM$^\#$} & 69.3 & 68.1 & 58.2 & 74.2 & 80.1 & 65.2 & \textbf{28.6} & 44.3 & 62.6 & 68.1 & \textbf{69.0} & 65.0 \\
  \hline
  \editmt & 59.3 & 62.5 & 34.7 & 69.8 & 68.0 & 50.6 & 12.1 & 38.0 & 31.2 & 52.3 & 45.6 & 51.8 \\
  $\quad$+ tag & 60.3 & 63.0 & 35.7 & 70.3 & 68.4 & 51.9 & 12.9 & 38.8 & 32.6 & 52.1 & 45.6 & 52.6 \\
  $\quad$+ R + tag & 56.0 & 53.9 & 45.9 & 64.9 & 68.5 & 50.0 & 17.7 & 39.7 & 44.9 & 59.9 & 52.8 & 53.3 \\
  $\quad$+ FT + tag & \textbf{70.6} & \textbf{71.5} & \textbf{57.8} & \textbf{78.2} & \textbf{82.0} & \textbf{66.2} & \textbf{28.0} & \textbf{45.1} & \textbf{61.1} & \textbf{67.7} & \textbf{66.8} & \textbf{66.0} \\
  $\quad$+ FT + R + tag & 66.4 & 63.6 & 57.3 & 71.3 & 77.6 & 60.5 & \textbf{28.0} & 42.1 & 60.9 & 66.7 & 65.0 & 62.3 \\
  \hline
  \editlevt & 59.5 & 62.6 & 34.8 & \textbf{69.0} & \textbf{66.9} & 54.2 & 12.2 & 37.8 & 31.0 & 49.0 & 44.8 & 52.0 \\
  $\quad$+ R & 49.6 & 49.1 & \textbf{43.6} & 57.8 & 61.0 & 42.9 & \textbf{16.5} & 34.3 & \textbf{41.0} & 49.2 & \textbf{50.6} & 47.6 \\
  $\quad$+ FT & \textbf{60.5} & \textbf{63.2} & 35.2 & 68.3 & 66.5 & \textbf{54.5} & 12.1 & \textbf{39.0} & 32.9 & \textbf{50.7} & 45.9 & \textbf{52.4} \\
  \hline
  \end{tabular}
  }
  \caption{BLEU scores for the multi-domain test sets. \textit{All} is computed by concatenating test sets from all domains, with $11$k sentences in total. \textit{Copy} refers to copying the similar translation in the output. \textit{+R} implies using the related segments instead of a full initial sentence for inference. Best performance in each block are in bold.\label{tab:fm-bleu}}
\end{table*}

\subsection{Experimental Settings\label{ssec:tm-settings}}

Our baselines reproduce two settings for TM-based MT: the \texttt{FM} setting of \citet{Bulte19fuzzy} and the \texttt{FM$^\#$} setting of \citet{Xu20boosting}. The former is trained using \textit{para} + \textit{similar} data, and the latter uses \textit{para} + \textit{related}. These two baselines are trained with the same configuration as in Section~\ref{ssec:data-bisynch}. We also report scores obtained by simply copying the retrieved similar translations, as in Section~\ref{ssec:bisynch-result}.

\editmt{} and \editlevt{} from Section~\ref{sec:bisynch} differ from \texttt{FM} and \texttt{FM$^\#$} both in task, and also in training domains. Hence, we consider fine-tuning (FT) our models. For \editmt{}, we use \textit{para}+\textit{similar}+ \textit{related} data and fine-tune for only $1$ epoch with a learning rate of $8e^{-5}$. As we do not have information about the edit operations required to change a similar translation into the reference, we set all editing tags as on for \textit{similar} data, and prefix the output with \texttt{[ins]} \texttt{[sub]} \texttt{[del]}. For the \textit{related} data, we conjecture that mostly insertions are needed as the irrelevant segments have already been removed. We thus only activate the insertion tag. For \editlevt, we only use \textit{similar}+\textit{related} and fine-tune for $1$ epoch with a learning rate of $9e^{-5}$. Our fine-tuned models can perform both translation with a similar sentence and with related segments.

\subsection{Results and Analysis\label{ssec:tm-results}}

\begin{table*}[!ht]
  \center
  \scalebox{0.85}{
  \begin{tabular}{l|cccccccccc|c}
  \hline
  Model \hfill $\Delta(\initrg,\trg)$ & 0 & 1 & 2 & 3 & 4 & 5 & 6 & 7 & 8-10 & >10 & All \\
  \hline
  $N$ & 540 & 2096 & 1107 & 882 & 827 & 782 & 689 & 607 & 1193 & 2277 & 11000 \\
  \hline
  copy & 100.0 & 82.3 & 74.2 & 67.2 & 62.1 & 51.7 & 50.5 & 48.2 & 40.7 & 33.8 & 52.6\\
  \hline
  \texttt{FM} & 91.6 & \textbf{93.3} & \textbf{86.5} & \textbf{82.3} & \textbf{79.2} & \textbf{70.5} & \textbf{69.0} & \textbf{68.0} & \textbf{60.9} & \textbf{49.5} & \textbf{67.3} \\
  \hline
  \editmt + tag & 95.3 & 80.8 & 72.9 & 68.0 & 62.7 & 52.4 & 50.7 & 49.6 & 41.5 & 34.3 & 52.6 \\
  $\quad$+ FT + tag & 91.6 & 91.1 & 85.8 & 80.9 & 77.7 & 68.8 & 68.4 & 66.6 & 59.0 & 48.0 & 66.0 \\
  \hline
  \editlevt & 94.5 & 79.5 & 72.1 & 66.3 & 61.0 & 49.9 & 49.6 & 47.5 & 40.5 & 33.7 & 52.0 \\
  $\quad$+ FT & \textbf{96.6} & 78.9 & 72.0 & 65.8 & 61.1 & 50.4 & 50.4 & 48.0 & 40.8 & 34.3 & 52.4 \\
  \hline
  \end{tabular}
  }
  \caption{BLEU scores for the multi-domain test sets broken down by the edit distance $\Delta$ between $\initrg$ and $\trg$. Each column represents a range of distances. $N$ denotes the number of sentences in each group.\label{tab:edit-dist-bleu-ter}}
\end{table*}

We evaluate with BLEU, and also show TER scores in Appendix~\ref{ssec:tm-more-result}. We reproduce in Table~\ref{tab:fm-bleu} the overall good performance of \texttt{FM} and \texttt{FM$^\#$}. Both significantly improve the initial similar translations. The generic \editmt{} performs much worse, and does not even match the copy results. When prefixed with the editing tag\footnote{We conjecture with a substitution tag for zero-shot inference as we have no information about the required edit type. Fine-tuned model uses the same tag as FT data.} (+tag), we observe small improvements (+$0.8$ BLEU on average), that are further increased in the \textit{related} scenario (+R). FT yields a much larger boost in performance (+$13.4$ BLEU). This highlights the effect of the task and domain mismatches on our initial results with \editmt. The \textit{related} setting also benefits from FT, albeit by a smaller margin (+$9$ BLEU). 

\begin{table}[!ht]
  \center
  \scalebox{0.85}{
  \begin{tabular}{l|cc}
  \hline
  Model & Office & ENV \\
  \hline
  copy & 54.7 & 59.6 \\
  \hline
  \texttt{FM} & 66.8 & 75.4 \\
  \texttt{FM$^\#$} & 64.0 & 70.6 \\
  \hline
  \editmt + tag & 56.2 & 60.3 \\
  $\quad$+ FT + tag & \textbf{68.6} & \textbf{78.6} \\
  \hline
  \editlevt & 53.0 & 60.0 \\
  $\quad$+ FT & 51.9 & 59.6 \\
  \hline
  \end{tabular}
  }
  \caption{BLEU scores on unseen domains.\label{tab:unseen}}
\end{table}

Our best results overall, using FT, are superior to \texttt{FM$^\#$} and close to that of \texttt{FM}. This has practical implications, since \texttt{FM$^\#$} and \texttt{FM} are specifically trained to transform a retrieved translation, whereas the generic \editmt{} is initially trained with artificial edits then only slightly fine-tuned on the in-domain data. To appreciate this difference, we evaluate our models on two unseen domains (Office and ENV), neither of which are used to train \texttt{FM} and \texttt{FM$^\#$}, nor to fine-tune \editmt{}. Results in Table~\ref{tab:unseen} unambiguously show that in this setting, the fine-tuned \editmt{} outperforms \texttt{FM}, suggesting that our edit-based model has not only adapted to the domain, but also to the task, as it can effectively perform zero-shot TM-based translations. Results obtained with the \editlevt{} model on these test sets lag far behind: even with FT on in-domain data, \editlevt{} still struggles to improve over the \textit{copy} baseline.

We also perform the analysis with a breakdown by the edit distance $\Delta$ as in Section~\ref{ssec:bisynch-result} on the multi-domain test sets. For the results in Table~\ref{tab:edit-dist-bleu-ter}, all $11$k test sentences are merged into one test set, then broken down by values of $\Delta$. The generic \editmt{} starts to improve the similar translation for $\Delta$$\geq$$3$. However, it is difficult for \editmt{} to detect very small changes ($\Delta$$<$$3$) without FT. Once fine-tuned, \editmt{} performs similar to \texttt{FM} for small changes, which further confirms that \editmt{} adapts to the TM-based translation task. \editlevt{} models, however, only slightly improve the copy baseline for similar translations requiring large changes ($\Delta$$\geq$$8$). However, it is better than other models at detecting $\initrg$s that do not need edits ($\Delta$$=$$0$). We also provide results broken down the merged test set by different edit operations in Appendix~\ref{ssec:tm-more-result}.

\section{Parallel Corpus Cleaning\label{sec:filtering}}

Our model restores synchronization between a pair of sentences. This is also useful for parallel TM cleaning tasks. Given a source sentence $\src$ and a possibly incorrect translation $\initrg$, we want to detect non-parallelism and \emph{to perform appropriate fixes}. We study how \editmt{} fares with this new problem on two publicly available datasets: first on the SemEval 2012\&3 Task 8: Cross-lingual Textual Entailment (CLTE, \citealp{Negri11divide,Negri12semeval,Negri13semeval}),\footnote{\url{https://ict.fbk.eu/clte-benchmark/}} then with the OpenSubtitles corpus \citep{Lison16opensubtitles}.

\subsection{Cross-lingual Textual Entailment\label{ssec:clte}}

The CLTE task aims to identify multi-directional entailment relationships between two sentences $\mathbf{x}_1$ and $\mathbf{x}_2$ written in different languages. We evaluate on the Fr-En direction, where $\mathbf{x}_1$ is in French and $\mathbf{x}_2$ is in English. The tagging mechanism of \editmt{} (see Section~\ref{ssec:model}) can readily be used for this classification task. Data descriptions and slight adjustments of tagging scheme are in Appendix~\ref{sec:filtering-appendix}.

We treat $\mathbf{x}_1$ as $\src$ and $\mathbf{x}_2$ as $\initrg$ to match the input format of \editmt, and perform zero-shot inference reusing the same \editmt{} model as in Section~\ref{sec:bisynch}. We concatenate $\mathbf{x}_1$ and $\mathbf{x}_2$ as input, and truncate the target sequence by only taking the first three edit tags as the predicted label for the corresponding input pair, treating \editmt{} as a mere classification (CLF) model. We also slightly fine-tune \editmt{} with the $500$ examples of CLTE training data for $5$ epochs with a learning rate of $8\mathrm{e}{-5}$.

Results are in Table~\ref{tab:clte}, together with the best scores reported in \citet{Negri13semeval} for years 2012 and 2013 and the scores reported in \citet{Carpuat17detecting}, which are the best reported performance we could find. Note that these scores may be quite weak, as pre-trained language models did not exist at that time. We have tried to apply a pre-trained XLM model \citep{Conneau19xlm} to report better baselines for CLTE. However, fine-tuning XLM with only \num{500} sentences for $5$ epochs did not outperform the reported baselines, as the fine-tuned XLM model needs to train new parameters for the linear output layer. Our \editmt, on the contrary, does not require any additional parameters during fine-tuning.

\begin{table}[!ht]
  \center
  \scalebox{0.85}{
  \begin{tabular}{l|cc}
    \hline
    Methods & 2012 & 2013 \\
    \hline
    Best SemEval13 & 0.570 & 0.458 \\
    \citet{Carpuat17detecting} & 0.604 &  0.436 \\
    \hline
    \editmt{} En-Fr & 0.350 & 0.284 \\
    \hfill + FT & \textbf{0.716} & 0.466 \\
    \hline
    \editmt{} Fr-En & 0.376 & 0.288 \\
    \hfill + FT & 0.710 & \textbf{0.530} \\
    \hline
  \end{tabular}
  }
  \caption{Accuracy scores on the SemEval CLTE tasks. FT denotes \editmt{} fine-tuned for classification.\label{tab:clte}} 
\end{table}

As can be seen in Table~\ref{tab:clte}, out-of-the-box \editmt{} fails to clearly detect the entailment relationships. This is not surprising, as there is a significant difference between our editing data and the CLTE test sets. For instance, the insertion initial translation $\initrg_{\operatorname{ins}}$ is always grammatically incorrect, while all sentences in CLTE are syntactically correct. However, after slight fine-tuning on the CLTE data, \editmt{} for both directions can quickly adapt to the task, achieving state-of-the-art performance. This again hints that \editmt{} actually learns to identify various cases of non-parallelism.

\subsection{Fixing OpenSubtitles Corpus\label{ssec:opensubtitles}}

We further evaluate the ability of \editmt{} to detect parallel sentences and fix noisy data. We experiment with the OpenSubtitles\footnote{\url{https://opus.nlpl.eu/OpenSubtitles-v2018.php}} data \citep{Lison16opensubtitles} for the En-Fr direction. In this corpus, the French side is translated from English, but with noisy segments. A standard approach is to filter out noisy sentences from the training data when building systems. We aim to study whether \editmt{} can automatically identify and edit, rather than discard, noisy sentence pairs, so that training can use the full set of parallel data. We measure performance on the \num{10159} segments of the En-Fr Microsoft Spoken Language Translation (MSLT) task \citep{Federmann16mslt}, which simulates a real world MT scenario.

The OpenSubtitles data is processed similarly as in Section~\ref{ssec:data-bisynch}. We first use the fine-tuned CLF model in the CLTE task to predict the relation for all sentence pairs in OpenSubtitles data. About $60\%$ of the data is classified as parallel, indicating that no edit operation is predicted for these segments. Models trained on the $60\%$ clean data are denoted as \texttt{filtered}. For the other $40\%$ presumably noisy data, we reuse the \editmt{} En-Fr model of Section~\ref{sec:bisynch} to fix the translations, using the predicted edit tag as a prefix on the target side (see Sections~\ref{sec:bisynch} and \ref{sec:MTwithTM}). Models trained on the edited data are noted as \texttt{fixed}. We train NMT models with either all data (\texttt{full}) or just the $40\%$ \texttt{noisy} data as baselines. For comparison, we also train a model using the same data size ($15.8$M) as the \texttt{noisy} subset, randomly selected from the \texttt{filtered} subset.

As shown in Table~\ref{tab:open}, aggressively filtering the noisy data improves over using the full training corpus (+$2$ BLEU for filtered) more than revising it (+$1$ BLEU for filtered + fixed). The second set of results yields similar conclusions with smaller datasets: here, the effect of automatically fixing a set of initially noisy data improves the BLEU score by $7.2$ points and closes half of the gap with a clean corpus of the same size. Note that these results are obtained without adaptation, simply reusing the pre-trained \editmt{} model of Section~\ref{sec:bisynch}. This suggests that in situations where the training data is small and noisy, editing-based strategies may provide an effective alternative to filtering.

\begin{table}[!ht]
  \center
  \scalebox{0.85}{
  \begin{tabular}{l|c|c}
  \hline
  Cleaning Method & BLEU & Corpus size \\
  \hline
  full & 44.7 & 41.6M \\
  filtered & \textbf{46.7} & 25.8M \\
  filtered + fixed & 45.7 & 41.6M \\
  \hline
  noisy & 32.2 & 15.8M \\
  fixed & 39.4 & 15.8M \\
  filtered (15.8M) & \textbf{46.7} & 15.8M \\
  \hline
  \end{tabular}
  }
  \caption{BLEU scores on MSLT taks of models trained with different subsets of OpenSubtitles.\label{tab:open}}
\end{table}

\section{Related Work\label{sec:related}}

The prediction of translations based on a source sentence and an initial translation is first explored in the context of IMT, using a left-to-right sentence completing framework proposed by \citet{Langlais00transtype}. The proposals of \citet{Green14human,Knowles16interactive,Santy19inmt} explore ways to generate translations based on given prefix hints. A more general setting, enabling arbitrary insertions thanks to LCD, is studied for online IMT systems in \citet{Huang21transmart}. Note that LCD was initially developed for other purposes, namely enforcing lexical or terminological constraints \citep{Hokamp17lexically,Post18fast,Hu19improved}. As this approach induces large decoding overheads, recent works in this thread explore NAT techniques: \citet{Susanto20lexically} propose to inject lexical constraints into an edit-based LevT \citep{Gu19levenshtein}, an approach improved by \citet{Xu21editor} with an additional repositioning operator. 

Recent attempts to revise initial translations are explored by \citet{Marie15touch}, who propose a touch-based scenario where users select usable translations segments, while the more questionable ones are automatically retranslated iteratively. This idea is revisited by \citet{Grangier18quickedit}, where undesired words in the initial translation are crossed-out. The authors use a dual source encoder to represent the initial translation along with the source sentence, which is also explored by \citet{Wang20touch} in a touch-editing case. The text infilling task is also considered by \citet{Xiao22bitiimt}, based on a single source encoder; see also \citet{Yang21wets} and \citet{Lee21intellicat} for related proposals. These studies consider a slightly different task than ours, as they only predict the missing part of the initial translation. Nevertheless, they can all be adapted to our generic \bisync{} scenario.

Similar approaches have also been studied in APE. Multi-source architectures have been explored in \textit{e.g.} \citep{Junczys-dowmunt18ms,Tebbifakhr18multisource,Shin18multiencoder,Pal18multisource}, whereas \citet{Hokamp17ensembling,Lopes19unbabel} jointly encode source and translation as one input. \citep{Wisniewski15predicting,Libovicky16cuni,Berard17lig} focus on learning edit operations. \citet{Junczys-dowmunt16loglinear} also propose to generate APE training data with round-trip translation.

\bisync{} also encompasses TM-based methods. \citet{Gu18search} use a second encoder to represent TM matches, an idea extended with a more compact representation of TM matches in \citet{Xia19graphbased}. As explained above, \citet{Bulte19fuzzy} use a single encoder, concatenating TM segments with the source. \citet{Xu20boosting} further add a second embedding feature indicating related segments in TM matches, and \citet{Pham20priming} propose to simultaneously consider the source and target sides of retrieved TMs. Retrieval-based MT is also explored in \citet{He21fast,Khandelwal21nearest,Cai21neural}, trying to make the performance gain less dependent on the quality of the retrieved TM matches, or to enforce a tighter coupling between TM matches and translations.

\section{Conclusion\label{sec:conclusion}}

This work introduced \bisync{}, the task of generating translations of a source sentence by editing a related target sentence. We have proposed various ways to create artificial initial translations for different editing types, that are needed for training. We have explored both autoregressive and non-autoregressive architectures, observing experimentally that our autoregressive \editmt{} model trained with artificial triplets performs bilingual resynchronization tasks in several real world scenarios. \editmt{} can also be quickly adapted to retrieval-based MT tasks, where we compared favorably to dedicated models. Finally, \editmt{} can also fix TMs by detecting parallel sentences and correct imperfect translations without adaptation. Another application that we wish to explore is APE.

Our NAT approach \editlevt{} is lagging behind \editmt. In the future, we would like to explore more NAT systems, which are computationally faster, and improve their performance. We intend to consider training curriculums and to modify the LevT model to better fit the \bisync{} task. We would also like to study ways to reduce the load of fully re-decoding the input sequence, especially when small changes, that need to be reproduced in the target, are iteratively applied to the source sentence.

\section*{Limitations}

The generation of editing data for each type of edit operations requires lots of efforts and resources. This requires two separately trained NMT models to generate the data for substitution via round-trip translation with LCD. The computational cost of LCD is very high compared to regular beam search, therefore consuming many computational resources. The generation of editing data for deletion also requires one separately trained model for each method with a complete decoding of the entire training corpora. Even though our data generation procedure is effective, the generation process may not be environmentally friendly. Due to computational limits, we were not able to conduct experiments on other languages pairs and tasks such as APE, in which large public datasets are available for other language pairs other than En-Fr.

As we decomposed the edits from one initial translation to the reference by basic edits (insertion, substitution and deletion), the generic \editmt{} and \editlevt{} models can mostly perform one type of edits at a time. It might be worth studying to combine several edit types into one single generated example, in order to approach more realistic scenarios for the generic models.

We mainly measured our results with BLEU and some additional scores in TER. However, other metrics like COMET \citep{Rei20comet} can also be interesting: as pointed out by \citet{Helcl22nonautoregressive}, BLEU may be less appropriate to measure valid translations than COMET for NAT models.

\section*{Acknowledgement}

We would like to thank the anonymous reviewers for their valuable suggestions and comments. This work was granted access to the HPC resources of IDRIS under the allocation 2022-[AD011011580R2] made by GENCI. The first author is partly funded by SYSTRAN and by a grant (Transwrite) from Région Ile-de-France.

\bibliography{biblio}
\bibliographystyle{acl_natbib}

\clearpage
\appendix

\section{Edit-based Model Configurations\label{sec:bisynch-appendix}}

We conduct experiments using \texttt{fairseq}\footnote{\url{https://github.com/pytorch/fairseq}} \citep{Ott19fairseq}. \editmt{} relies on the Transformer-base model of \citet{Vaswani17attention}. We use a hidden size of $512$ and a feedforward size of \num{2048}. We optimize with Adam with a maximum learning rate of $0.0007$, an inverse square root decay schedule, and \num{4000} warmup steps. We also tie all input and output embedding matrices \citep{Press17using,Inan17tying}. \editmt{} is trained with mixed precision and a batch size of \num{8192} tokens on 4 V100 GPUs for $300$k iterations. We save checkpoints for every \num{3000} iterations and average the last $10$ saved checkpoints for inference. For \editlevt, we follow \citet{Gu19levenshtein}, using a maximum learning rate of $0.0005$ with \num{10000} warmup steps and a larger batch size of \num{16384}. For inference, we set a maximum decoding round of $10$.

\section{Details for MT with TMs\label{sec:tm-appendix}}

\subsection{Datasets and Processing Details\label{ssec:tm-data-detail}}

\begin{table}[!ht]
  \center
  \scalebox{0.85}{
  \begin{tabular}{l|rrr}
  \hline
  Domain & Train & FM ratio & FM train \\
  \hline
  ECB & \num{195956} &   51.73\% & \num{234943} \\
  EMEA & \num{373235} &  65.68\% & \num{624109} \\
  Epps & \num{2009489} & 10.12\% & \num{465228} \\
  GNOME & \num{55391} &  39.31\% & \num{42697} \\
  JRC & \num{503437} &   50.87\% & \num{587859} \\
  KDE & \num{180254} &   36.00\% & \num{136456} \\
  News & \num{151423} &   2.12\% & \num{4048} \\
  PHP & \num{16020} &    34.93\% & \num{10350} \\
  TED & \num{159248} &   11.90\% & \num{39895} \\
  Ubuntu & \num{9314} &  20.32\% & \num{1738} \\
  Wiki & \num{803704} &  19.87\% & \num{409755} \\
  \hline
  Total & \num{4457471} & 24.27\% & \num{2557078} \\
  \hline
  Office & \num{49845} & 43.76\% & - \\
  ENV & \num{13632} & 6.81\% & - \\
  \hline
  \end{tabular}
  }
  \caption{Data used for experiments of Section~\ref{ssec:tm-settings}. \textit{FM ratio} is the ratio of sentences with at least one matched similar translations, \textit{FM train} is the actual number of examples augmented with a similar translation used for training, after setting aside \num{1000} test sentences for each domain. Each training sentence is matched with up to $3$ similar translations.\label{tab:data-stat}}
\end{table}

Our experiments use the same multi-domain corpus as \citet{Xu20boosting}. This corpus contains $11$ different domains for the En-Fr direction, collected from OPUS\footnote{\url{https://opus.nlpl.eu/}} \citep{Tiedemann12parallel}: documents from the European Central Bank (ECB); from the European Medicines Agency (EMEA); Proceedings of the European Parliament (Epps); legislative texts of the European Union (JRC); News Commentaries (News); TED talk subtitles (TED); parallel sentences extracted from Wikipedia (Wiki); localization files (GNOME, KDE and Ubuntu) and manuals (PHP). All these data were deduplicated prior to training. To evaluate the ability of our models to actually make use of TMs instead of memorizing training examples, we also test on two unseen domains: OpenOffice from OPUS and the PANACEA environment corpus\footnote{\url{http://catalog.elda.org/en-us/repository/browse/ELRA-W0057/}} (ENV). We follow \citet{Xu20boosting} and search for the top $3$ similar translations based on Fuzzy Match with a similarity score greater than $0.6$ as computed by Equation~\eqref{eq:similarity} on the source side and without an exact match. Note that the ratio of sentences with at least one similar translation greatly varies across domains, as shown in Table~\ref{tab:data-stat}. When reproducing the \textit{related} setting, \citet{Xu20boosting} used a placeholder token to mark the positions where segments are deleted. As discussed in Section~\ref{ssec:data-generation}, our models do not include such information. Therefore, we do not use placeholders for the \textit{related} data. 

\begin{table*}[!ht]
  \center
  \scalebox{0.8}{
  \begin{tabular}{l|ccccccccccc|c}
  \hline
  TER $\downarrow$ & ECB & EMEA & Epps & GNOME & JRC & KDE & News & PHP & TED & Ubuntu & Wiki & All \\
  \hline
  copy & 0.435 & 0.377 & 0.659 & 0.263 & 0.294 & 0.457 & 0.999 & 0.703 & 0.653 & 0.372 & 0.502 & 0.488 \\
  \hline
  \texttt{FM} & \textbf{0.286} & \textbf{0.301} & 0.374 & \textbf{0.160} & \textbf{0.138} & 0.312 & 0.642 & \textbf{0.572} & 0.328 & \textbf{0.227} & \textbf{0.306} & \textbf{0.314} \\
  \texttt{FM$^\#$} & 0.299 & 0.332 & \textbf{0.365} & 0.202 & 0.152 & \textbf{0.309} & \textbf{0.634} & 0.605 & \textbf{0.327} & 0.234 & 0.309 & 0.327 \\
  \hline
  \editmt & 0.418 & 0.390 & 0.641 & 0.260 & 0.285 & 0.455 & 0.904 & 0.646 & 0.643 & 0.359 & 0.520 & 0.472 \\
  $\quad$+ tag & 0.412 & 0.382 & 0.638 & 0.256 & 0.283 & 0.452 & 0.899 & 0.643 & 0.635 & 0.360 & 0.520 & 0.468 \\
  $\quad$+ R + tag & 0.397 & 0.422 & 0.508 & 0.268 & 0.237 & 0.428 & 0.825 & 0.636 & 0.520 & 0.282 & 0.457 & 0.430 \\
  $\quad$+ FT + tag & \textbf{0.300} & \textbf{0.308} & 0.376 & \textbf{0.182} & \textbf{0.146} & \textbf{0.314} & 0.643 & \textbf{0.595} & 0.351 & \textbf{0.237} & \textbf{0.329} & \textbf{0.326} \\
  $\quad$+ FT + R + tag & 0.322 & 0.357 & \textbf{0.375} & 0.219 & 0.168 & 0.342 & \textbf{0.640} & 0.623 & \textbf{0.350} & 0.238 & 0.338 & 0.346 \\
  \hline
  \editlevt & 0.437 & \textbf{0.390} & 0.655 & \textbf{0.267} & 0.296 & \textbf{0.463} & 0.996 & 0.720 & 0.656 & 0.399 & 0.528 & 0.496 \\
  $\quad$+ R & \textbf{0.434} & 0.446 & \textbf{0.503} & 0.314 & \textbf{0.294} & 0.471 & \textbf{0.780} & \textbf{0.649} & \textbf{0.515} & \textbf{0.370} & \textbf{0.478} & \textbf{0.458} \\
  $\quad$+ FT & 0.443 & \textbf{0.390} & 0.663 & 0.276 & 0.306 & 0.467 & 1.008 & 0.722 & 0.677 & 0.389 & 0.531 & 0.502 \\
  \hline
  \end{tabular}
  }
  \caption{TER scores on multi-domain test sets. \textit{All} is computed by concatenating test sets from all domains, with $11$k sentences in total. \textit{Copy} refers to copying the similar translation in the output. \textit{+R} implies using the related segments instead of a full initial sentence for inference. Best performance in each block are in bold.\label{tab:fm-ter}}
\end{table*}

\begin{table*}[!ht]
  \center
  \scalebox{0.85}{
  \begin{tabular}{l|cccccccccc}
  \hline
  BLEU$\uparrow$ & = & Ins & Sub & Del & Ins+Sub & Ins+Del & Sub+Del & Ins+Sub+Del \\
  \hline
  $N$ & 540 & 316 & 3260 & 316 & 2865 & 58 & 2634 & 1011 \\
  \hline
  copy & 100.0 & 72.0 & 67.9 & 75.4 & 32.5 & 69.8 & 34.0 & 47.3 \\
  \hline
  \texttt{FM} & 91.6 & \textbf{80.6} & \textbf{86.6} & 82.9 & \textbf{50.0} & 67.4 & \textbf{58.4} & \textbf{63.0} \\
  \hline
  \editmt + tag & 95.3 & 75.7 & 67.0 & 77.2 & 34.2 & 68.5 & 37.4 & 48.0 \\
  $\quad$+ FT + tag & 91.6 & 79.7 & 84.6 & \textbf{85.8} & 48.3 & 69.9 & 57.6 & 60.8 \\
  \hline
  \editlevt & 94.5 & 75.3 & 65.9 & 73.3 & 33.1 & 69.4 & 33.7 & 46.9 \\
  $\quad$+ FT & \textbf{96.6} & 75.5 & 65.6 & 74.1 & 34.3 & \textbf{70.1} & 33.9 & 47.0 \\
  \hline
  \end{tabular}
  }
  \caption{BLEU scores for the multi-domain test sets broken down by the edit operations between $\initrg$ and $\trg$. Each column represents a combination of edits. $N$ denotes the number of sentences in each group.\label{tab:edit-type-bleu-ter}}
\end{table*}

\section{Additional Results\label{ssec:tm-more-result}}

Table~\ref{tab:fm-ter} reports results on multi-domain test set for the task of translating with TMs measured on TER using SacreBLEU \citep{Post18sacrebleu}. TER results show that even the generic \editmt{} model actually identifies useful edits, as we see improvements with respect to the copy baseline. 

Table~\ref{tab:edit-type-bleu-ter} shows results on BLEU broken down the aggregate test set by edit operations. We observe that the generic Edit-MT struggles to perform well when substitutions are needed. FT vastly improves the ability to substitute and delete from $\initrg$, separately or even in combination. Fine-tuned Edit-MT even outperforms \texttt{FM} when only deletions are required. 

\section{Details for Parallel Corpus Cleaning\label{sec:filtering-appendix}}

In the CLTE task, the goal is to identify multi-directional entailment relationships between two sentences $\mathbf{x}_1$ and $\mathbf{x}_2$, written in different languages. Each ($\mathbf{x}_1, \mathbf{x}_2$) pair in the dataset is annotated with one of the following relations: \textit{Bidirectional} ($\mathbf{x}_1$$\Leftrightarrow$$\mathbf{x}_2$): the two fragments entail each other (semantic equivalence); \textit{Forward} ($\mathbf{x}_1$$\Rightarrow$$\mathbf{x}_2$ $\&$ $\mathbf{x}_1$$\nLeftarrow$$\mathbf{x}_2$): unidirectional entailment from $\mathbf{x}_1$ to $\mathbf{x}_2$; \textit{Backward} ($\mathbf{x}_1$$\nRightarrow$$\mathbf{x}_2$ $\&$ $\mathbf{x}_1$$\Leftarrow$$\mathbf{x}_2$): unidirectional entailment from $\mathbf{x}_2$ to $\mathbf{x}_1$; \textit{No Entailment} ($\mathbf{x}_1$$\nLeftrightarrow$$\mathbf{x}_2$): no entailment between $\mathbf{x}_1$ and $\mathbf{x}_2$. The dataset contains a training set of $500$ pairs, and two test sets of the same size (test-2012 and test-2013). 

\begin{table}[!ht]
  \center
  \scalebox{0.85}{
  \begin{tabular}{l|cc}
  \hline
  CLTE Fr-En & \editmt{} En-Fr &\editmt{} Fr-En \\
  \hline
  Bidirectional   & Copy & Copy \\
  Forward         & Deletion & Insertion \\
  Backward       & Insertion & Deletion \\
  No Entailment & Substitution & Substitution \\
  \hline
  \end{tabular}
  }
  \caption{Label conversion scheme between CLTE task and \editmt{} editing tags. \label{tab:label}}
\end{table}

The tagging mechanism of Edit-MT described in Section~\ref{ssec:model} can readily be used for this classification task with slight adjustments represented in Table~\ref{tab:label}.

We have not performed hyperparameter searching for FT, even though carefully fine-tuned models may achieve even better performance. 

\end{document}